\documentclass[letterpaper, 10 pt, conference]{ieeeconf}  

\IEEEoverridecommandlockouts                              

\usepackage{graphics} 
\usepackage{cite}
\usepackage{amsmath,amssymb,amsfonts}
\usepackage{algorithmic}
\usepackage{graphicx}
\usepackage{textcomp}
\usepackage{xcolor}
\title{\LARGE \bf
Learning manipulation of steep granular slopes \\for fast Mini Rover turning
}

\author{Deniz Kerimoglu$^{1,*}$, Daniel Soto$^{1,*}$, Malone Hemsley$^{2}$, Joseph Brunner$^{1}$, Sehoon Ha$^{3}$, Tingnan Zhang$^{4}$ \\and Daniel I. Goldman$^{1}$
\thanks{*Co-first authors}
\thanks{$^{1}$Deniz Kerimoglu, Daniel Soto, Joseph Brunner and Daniel I. Goldman are with School of Physics,
        Georgia Institute of Technology, Atlanta, USA
        {\tt\small dkerimoglu6@gatech.edu}}%
\thanks{$^{2}$Malone Hemsley is with the Department of Applied Physics and Engineering, Morehouse College,
        Atlanta, USA}%
\thanks{$^{3}$Sehoon Ha is with the School of Interactive Computing, Georgia Institute of Technology,
        Atlanta, USA}%
\thanks{$^{4}$Tingnan Zhang is with the Google Robotics, Mountain View, California, USA}%
}

\begin{document}

\maketitle
\thispagestyle{empty}
\pagestyle{empty}

\begin{abstract}
Future planetary exploration missions will require reaching challenging regions such as craters and steep slopes. Such regions are ubiquitous and present science-rich targets potentially containing information regarding the planet’s internal structure. Steep slopes consisting of low-cohesion regolith are prone to flow downward under small disturbances, making it very challenging for autonomous rovers to traverse. Moreover, the navigation trajectories of rovers are heavily limited by the terrain topology and future systems will need to maneuver on flowable surfaces without getting trapped, allowing them to further expand their reach and increase mission efficiency. 

In this work, we used a laboratory-scale rover robot and performed maneuvering experiments on a steep granular slope of poppy seeds to explore the rover's turning capabilities. The rover is capable of lifting, sweeping, and spinning its wheels, allowing it to execute leg-like gait patterns. The high-dimensional actuation capabilities of the rover facilitate effective manipulation of the underlying granular surface. We used Bayesian Optimization (BO) to gain insight into successful turning gaits in high dimensional search space and found strategies such as differential wheel spinning and pivoting around a single sweeping wheel. We then used these insights to further fine-tune the turning gait, enabling the rover to turn 90 degrees at just above 4 seconds with minimal slip. Combining gait optimization and human-tuning approaches, we found that fast turning is empowered by creating anisotropic torques with the sweeping wheel.
\end{abstract}

\section{INTRODUCTION}
Almost all portions of the celestial bodies and a significant portion of the granular surface on Earth remain inaccessible to the majority of wheeled and tracked vehicles. This is partly due to the limitations of locomoting robots that struggle to effectively interact with their immediate surroundings. Particularly, wheeled systems can get stuck in loose granular media as they sink deeper into the material during the wheel-spinning motion \cite{Schaefer2010}. These difficulties intensify in the vicinity of craters and other destinations characterized by steep, flowable surfaces, where a minimal disturbance could potentially cause yielding and downhill flow. Such regions are scientifically valuable locations, potentially harboring layers of geological history as well as containing shaded areas where water ice might be preserved \cite{sharpton2014outcrops,mahanti2018small}. To explore these difficult terrains, the next generation of exploratory robots will need to navigate over steep flowable slopes. 

Future missions will also require autonomous systems to negotiate difficult terrain topologies by performing effective maneuvers to increase access to a diverse range of terrain types. This will allow currently-denied extreme terrain to become traversable, hence achieving scientific mission goals would be less time- and energy-consuming due to the greatly shortened paths.

Recent research has addressed some of the mobility limitations to some extent. For loose regolith/soil on level ground, refined terramechanical models \cite{meirion2011modified} and new methods \cite{li2013terradynamics} based on granular resistive force theory \cite{reid2016actively} better predict the rover-terrain interaction. On the robotics side, various wheeled and non-wheel robotic platforms have been developed \cite{thoesen2021planetary,sanguino201750,weibel2023towards}. Few have achieved robust mobility in loose granular slopes, \cite{inotsume2016finding,lyu2023modeling,arm2023scientific}. However, robotic turning has not been extensively explored on steep loose granular slopes yet. 

Our previous works explored the climbing gaits of a laboratory-scale rover robot, namely Mini Rover, a scaled-down version of a robotic lunar rover prototype called NASA Resource Prospector 15 \cite{shrivastava2020material,karsai2022real}. The Mini Rover incorporates wheel and leg-like appendages that actively manipulate flowable terrain to climb granular slopes \cite{shrivastava2020material}. We extended Mini Rover's capabilities in \cite{karsai2022real} using Bayesian optimization, enabling the discovery of fast climb gaits by exploiting the fluid and solid aspects of granular materials. These investigations exemplify a wider category of interactions involving locomotor and flowable terrain. The locomotor becomes integrated with its surroundings behaving as a unified structure and actively modifies its environment in order to achieve success.  
In this work, motivated by leveraging terrain manipulation strategies, we study the turning of the Mini Rover robot on a steep granular slope. To this end, simulation environments to investigate the Mini Rover gaits are impractical since simulating interactions between the robot and the millions of granular particles is extremely time-consuming even with the use of high performance computing tools. Moreover, both the robot and the terrain exhibit complex, nonlinear dynamics, hence deriving closed-form solutions to represent such interactions is not yet possible, hindering model-based simulation approaches. Therefore we rely on laboratory experiments and optimization methods to find and investigate Rover gait patterns. 

Inspired by recent progress in robot learning we used Bayesian optimization \cite{garnett2023bayesian} to search the extensive gait space and acquire gait cues. We started by picking a successful gait from our previous work \cite{karsai2022real} which was optimized to rapidly climb granular inclines and modified it to prompt turning behavior in a desired direction. 
This gait is then used as a seed for the Bayesian optimizer which learned faster turning strategies. 
The insights from the optimization are then used to understand the locomotor-terrain interactions and to perform targeted systematic experiments for further gait development. 

We discovered that the Mini Rover achieves fast turning by combining multiple terrain manipulation strategies. Mini Rover first creates a yaw motion via differential wheel spinning, followed by a single rear wheel sweeping motion that exploits the multiphase feature of the granular media by generating anisotropic torques. Finally, the rover aligns its roll angle to the changing slope condition by extending and retracting its wheels accordingly. Hence, our ML-inspired new gait outperforms gaits found solely by using Bayesian Optimization and manual tuning by a large margin, turning in 4.5 seconds as opposed to 2 minutes, respectively. 

\section{Materials and Methods}
\subsection{Robotic Platform: Mini Rover}
\begin{figure}[t]
\centerline{\includegraphics[width=0.46\textwidth]{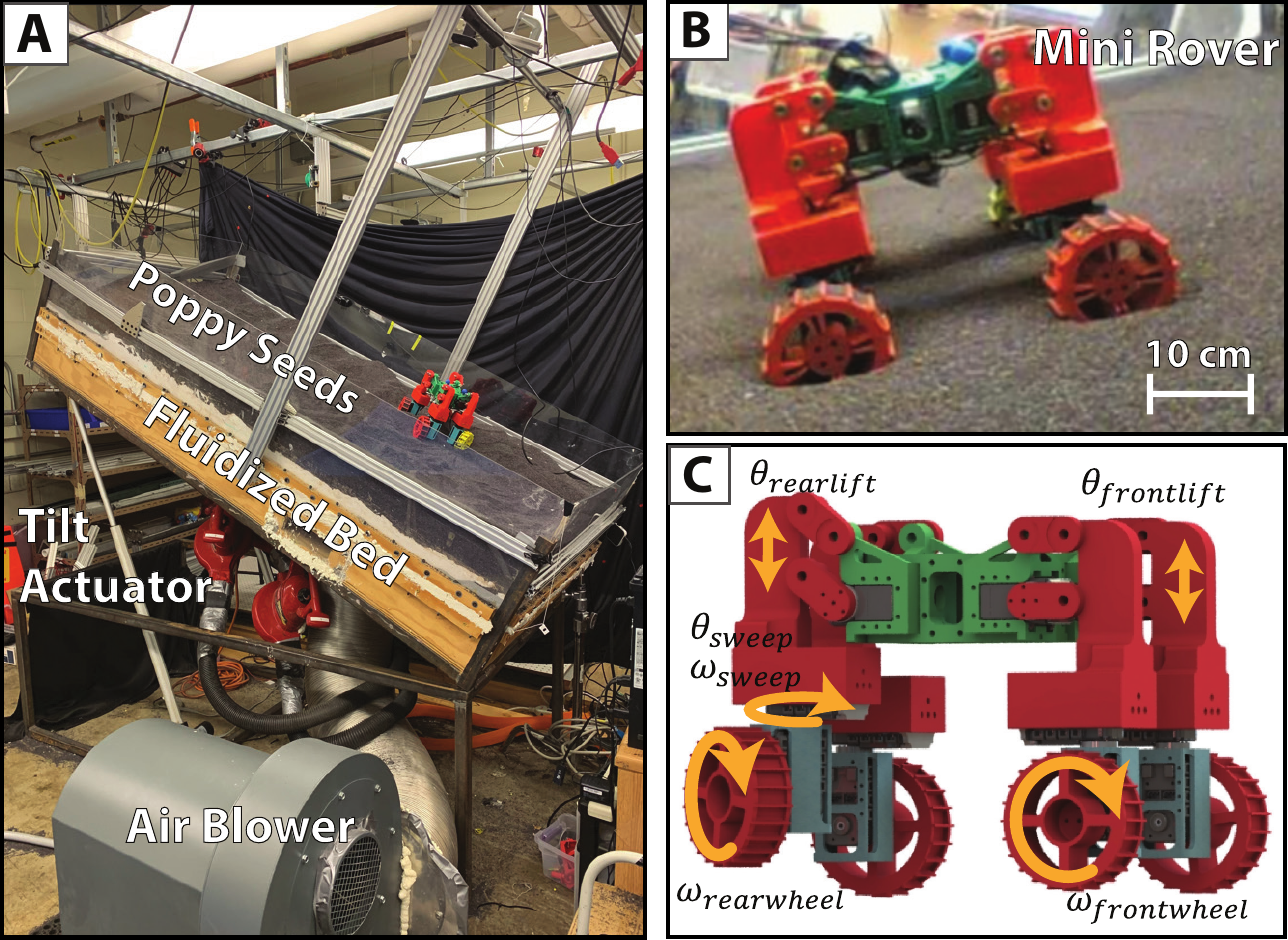}}
\caption{Experimental apparatus and the rover images. (A) shows the tiltable and air-fluidized testbed of poppy seeds. (B) shows the Mini Rover on loose granular slop. (C) illustrates the actuation parameters of the Mini Rover.}
\label{fig1_Rover}
\end{figure}

The Mini Rover consists of 12 Dynamixel AX-12 actuators and 3D-printed linkages and grousers (Fig. \ref{fig1_Rover}B). Each wheel-leg complex can lift up and down as well as spin and sweep. The wheel-leg appendages function in a way that resembles legs for propulsion, but they also spin to disrupt the surrounding terrain as wheels operate. The rover's main body is revised to be lighter to reduce the effect of inertial forces during fast turning.

The experimental testbed that we run the Mini Rover is a tiltable bed containing 1 mm diameter poppy seeds (Fig. \ref{fig1_Rover}A). The bed size is 2.5 m by 1.2 m, which is large enough to avoid boundary effects during turning. The poppy seeds represent a flowable and deformable terrain, a laboratory analog for flowable and deformable surfaces \cite{shrivastava2020material}. The granular terrain is reset to a loosely consolidated state after each experiment via air fluidization, creating a controlled uniform initial state for systematic experiments (Fig. \ref{fig1_Rover}A). For robot tracking purposes, we placed three blue spherical markers on top of the rover and three webcams were mounted on the testbed to capture the top and side-view videos. Markers constitute a triangle which is used to track the position and orientation of the rover using MATLAB Computer Vision Toolbox. Control of the rover is achieved by creating open loop gaits trajectories in MATLAB which are tracked internally via Dynamixel actuators' motor drivers and PID controllers. 

To understand the locomotor-terrain interaction we performed experiments to measure the torque generated by the wheel moving in poppy seeds. The single-wheel sweep and spin experiments were performed on a platform capable of moving vertically via a Firgelli linear actuator. A 6-axis Force sensor from ATI Industries is mounted to the linear actuator. The sweeping and spinning wheel system is mounted to the force sensor to capture the resistive torques induced by the poppy seeds. First, the wheel system is immersed 2 cm into the poppy seeds, and then the substrate is air fluidized, resetting the media in the vicinity of the wheel. Then the wheel started to spin and sweep at low and high rates. The resistive torque data is logged to a desktop computer via a NI-DAQ card and Labview software. 

\subsection{BO-RRP Gait to TRRP Gait}
\begin{figure}[t]
\centerline{\includegraphics[width=0.46\textwidth]{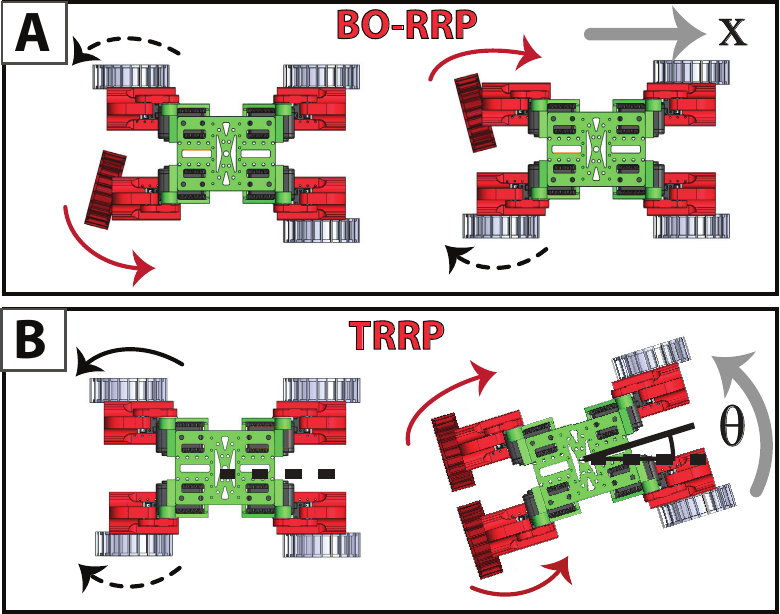}}
\caption{Mini Rover gait diagrams. (A) illustrates the snapshots of the rover executing BO-RRP. The white rear wheel sweeps-out 100° (toward the rear of the rover), while the red wheel sweeps-in, resetting the wheel position (toward the front of the rover). The sweep-out direction is shown in black arrows. The dashed arrow indicates that the wheel spinning is disabled during sweep-out. (B) shows the TRRP gait, where both rear wheels perform sweeping-out at the same time while the right rear wheel spinning is disabled. Once sweep-out is completed, both rear wheels spin and sweep-in.}
\label{Fig2_GaitDiag}
\end{figure}

To start inducing turning motion with the Mini Rover we modified the Bayesian Optimized Rear Rotator Pedaling (BO-RRP) gait, a previously studied successful gait pattern in \cite{karsai2022real} (see (Fig. \ref{Fig2_GaitDiag}A)). BO-RRP lifts, spins, and sweeps the rear wheel-leg appendages in an alternating style, exploiting the multi-phase feature of granular media by selectively solidifying/fluidizing the mound around the wheels. In the "solidifying phase" the wheel spin halts and the sweeping wheel exerts high pressure that pushes grains backward, creating high reaction force, as shown with dashed black curves in Fig. \ref{Fig2_GaitDiag}A. In the "fluidizing phase", the wheel spin is activated to create fluidizing shear when resetting the appendage's position, generating a low reaction force, as shown with dashed red curves. Although this strategy significantly increased the rover's locomotion speed, it was incapable of creating maneuvering gaits and hence would only climb straight. 

We modified BO-RRP by disabling the selective solidification/fluidization of the rear wheel, i.e. disabling the spinning motion while sweeping the wheel toward the rear of the rover (sweep-out), on the turning side of the Mini Rover (dashed black arrow in Fig. \ref{Fig2_GaitDiag}B). In addition, we set RRPs of both sides to run simultaneously, similar to breaststroke motion. In doing so the modified gait, called TRRP, created a higher push force on one side and a lower force on the other, yielding a differential thrust and eventually turning in one direction.
\subsection{Bayesian Optimization of TRRP}
We designed the new TRRP gait based on our prior experience, however, these changes may cause the parameters inherited from BO-RRP to be suboptimal. Therefore, we take another round of Bayesian optimization to optimize control parameters further, a technique that has demonstrated effectiveness in handling intricate robot-terrain interactions \cite{ha2020learning,yu2020learning}. Operating efficiently on systems with low sampling rates, BO treats the objective function as a stochastic function without making any assumptions by assigning a prior probability distribution using a Gaussian process. Over multiple iterations, the outcomes of function evaluations update the priors to create a posterior distribution, continuing until converging to the maximum of the objective function. 

The Bayesian optimization is initiated with TRRP so that the gait parameter search starts around a relatively successful gait. The Bayesian optimization is performed for 30 experiment episodes on a 25$^\circ$ granular slope and we omitted the results ending in failure. The open-loop gaits are set to run for 60 cycles taking about 2 minutes. The turning trials are initiated at a body orientation of 0$^\circ$ and the objective function is set to be 90$^\circ$ in the counterclockwise direction with respect to the world frame. We tracked the body orientation, and after each experiment, we updated the objective function to be the achieved turning angle. BO primarily varied the sweep range of the rear wheels, the spin direction, and the speed of all wheels as well as the extension/retraction range of wheel-leg appendages. The optimization search space is illustrated in Fig. \ref{fig1_Rover}C. 

\section{Results}

\subsection{Baseline Gaits}
We start maneuvering experiments with the TRRP gait. TRRP achieved gradual turning, taking over 2.5 minutes to get to 90$^\circ$ of body orientation (see Fig. \ref{Fig4_PerfEval} blue curve). Moreover, the Mini Rover drifted and slid down moderately during turning and failed 2 times out of 7 trials. This is in part due to its inability to manipulate the underlying terrain to create enough turning thrust as well as limitations in its capacity to adapt to the changes in the slope.

\subsection{Optimized TRRP Gait (BO-TRRP)}
\begin{figure}[t]
\centerline{\includegraphics[width=0.46\textwidth]{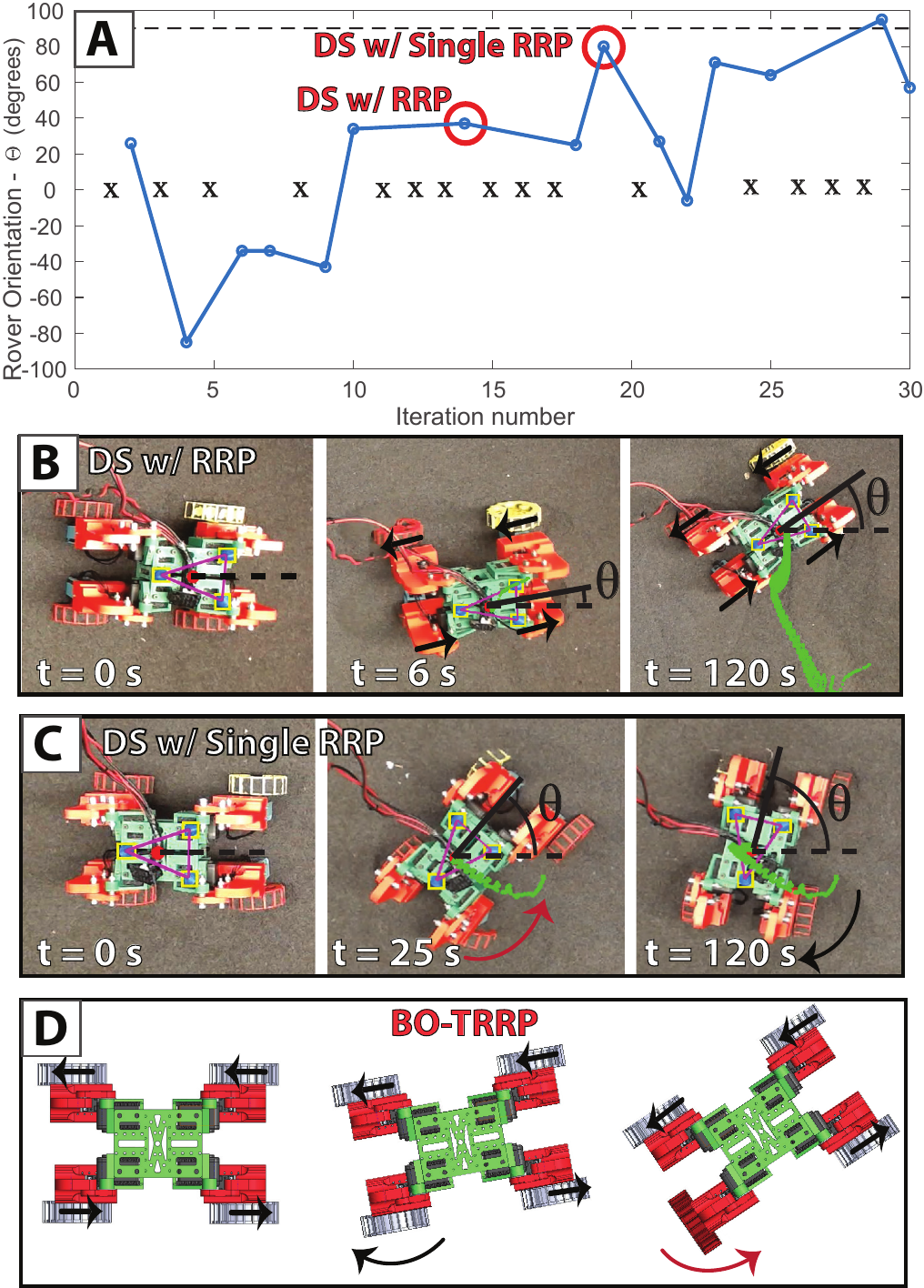}}
\caption{Evaluation of Bayesian Optimization of turning gaits. (A) depicts the evolution of Mini Rover's body orientation as a function optimization iteration. The circles and crosses represent successful and failed experiments, respectively. (B) and (C) illustrate the tracked images of the Mini Rover's CoM (green curves) and orientation ($\theta$) while turning on 25$^\circ$ slope at iterations 14 and 19, respectively. The optimization iterations 14 and 19 discovered DS and single RRP gaits, respectively. (D) The snapshot images of the rover executing the best-performing optimized gait (BO-TRRP). The straight and curved arrows indicate the wheel's spinning and sweeping directions whereas the straight arrows on the wheels depict the direction of the wheel spinning, respectively. The white and red rear wheels indicate sweeps-out (toward the rear of the rover) and sweeps-in (toward the front of the rover), respectively. BO-TRRP starts with the DS motion which is followed by the single RRP of the right rear wheel. The cooperation of these two manipulation strategies induces rover turning.}
\label{Fig3_ML_Eval}
\end{figure}

Next, we fed the TRRP gait parameters into Bayesian Optimization and evaluated the turning performance of the Mini Rover with iterative experiments, with a random exploration number of 4. Illustrated in Fig. \ref{Fig3_ML_Eval}A, in the first 10 iterations, the rover turned in the clockwise direction, opposite to the desired orientation, peaking its performance at iteration 4. 
Despite being in the wrong direction, the Mini Rover achieved turning $82^{\circ}$ by discovering the tank-turning maneuver with the front wheels spinning in the reverse direction to each other, a conventional turn-in-place strategy. In doing so, the Mini Rover turns in place via the torque generated by the reciprocal spinning of the front wheels. The Mini Rover started to turn in the right direction at iteration 10. Then, at iteration 14 the optimizer learned the differential spinning (DS) of both the front and rear wheels and turned in the desired orientation as illustrated with the time-lapse images in Fig. \ref{Fig3_ML_Eval}B. The Mini Rover still performed RRP motion with the rear wheels, however, incorporating DS motion enabled moderate turning. Consequently, at iteration 19 the optimizer learned to perform RRP only with the appendage located at the opposing side of the desired orientation (single RRP) as shown in Fig. \ref{Fig3_ML_Eval}C, while the remaining wheels kept performing DS motion. The cooperation of both strategies, called BO-TRRP, resulted in almost 90$^\circ$ of turning as shown with the gait diagram in Fig. \ref{Fig3_ML_Eval}D. 

\begin{figure}[b]
\centerline{\includegraphics[width=0.46\textwidth]{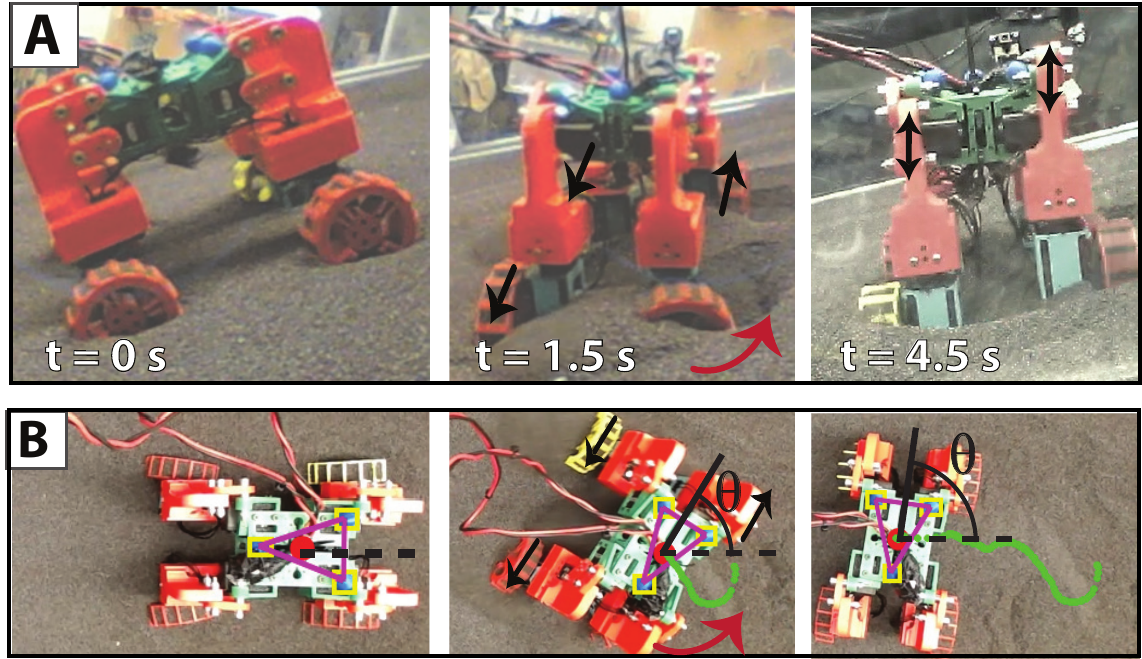}}
\caption{Mini Rover executing ML-inspired gait. (A) and (B) illustrate the side and top view of the Mini Rover performing fast turning with ML-inspired gait on a 25$^\circ$ slope. Straight and curved arrows depict spin and sweep directions, respectively. Double-arrows illustrate the difference in the appendage leg extension and retraction.}
\label{Fig6_BestGait}
\end{figure}
\begin{figure}[t]
\centerline{\includegraphics[width=0.46\textwidth]{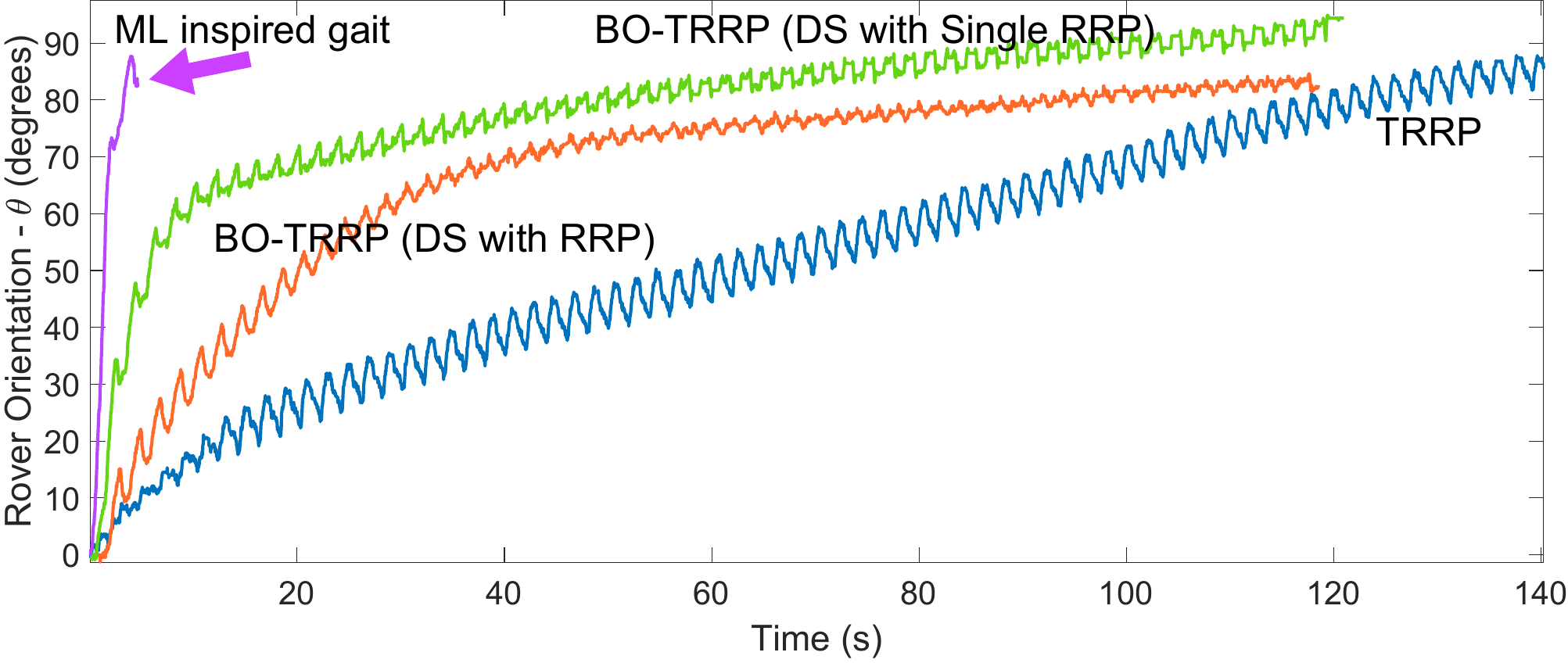}}
\caption{Time evolution of Mini Rover's orientation using various turning gaits on a slope of 25$^\circ$. TRRP gait achieves turning 90$^\circ$ gradually in about 2.5 minutes. Optimized gaits start turning fast initially, however, their performance saturates as they continue turning. ML-inspired gait facilitates effective terrain manipulation strategies to turn 90$^\circ$ in just above 4 seconds.}
\label{Fig4_PerfEval}
\end{figure}
\begin{figure}[t]
\centerline{\includegraphics[width=0.46\textwidth]{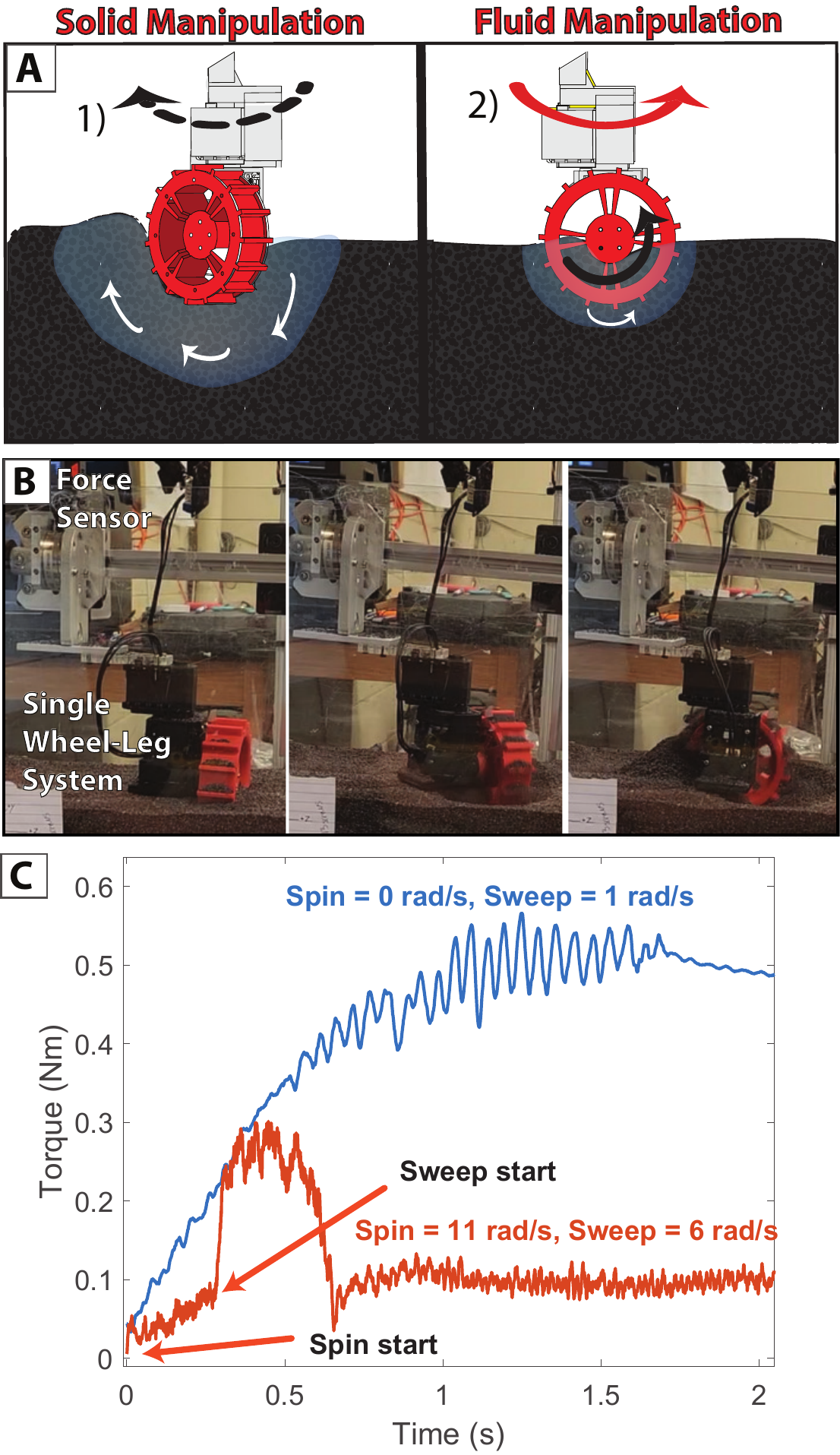}}
\caption{Single wheel spin and sweep experiments. (A) Illustration of solid and fluid manipulation of granular terrain with the single wheel-leg system. In (1) the wheel stops spinning while sweeping at a slow speed, creating more resistive torque. In (2) the wheel spins and sweeps at a high rate, leading to less resistive torque. (B) The apparatus used to perform spin and sweep experiments. (C) The time evaluation of torques obtained from the experiments. The blue curve depicts the net torque of a non-spinning wheel sweeping 90$^\circ$ at 1 rad/s. The torque gradually increases and saturates at 1 second. The orange curve represents the wheel spinning and sweeping 90$^\circ$ at 11 rad/sand and 6 rad/s, respectively. The wheel spin yields linear-like torque where the onset of sweep generates a peak in the torque.}
\label{Fig5_SoloWheel}
\end{figure}
Continued experiments showed little progress in discovering novel gait patterns, only yielding faster DS motion at iteration 29. The plateauing of gait performance is an indication of entrapment into a local minimum, a commonly observed phenomenon in high-dimensional systems. Additionally, despite the fact that BO provided us with important strategies for rover turning, the process was rather slow taking around 2 minutes. 

The high-dimensional actuation scheme empowers the Mini Rover to effectively manipulate the deformable terrain that it stands on. However, the search for optimized maneuvering gaits in the vast gait space by only using hardware experiments is challenging and time-consuming. To facilitate faster turning and avoid local minima, we analyzed instances of optimized gaits. 
In doing so, we aim to understand how the locomotor-terrain interaction induces turning and design more effective gaits. Closer investigation of the BO-TRRP revealed that Mini Rover significantly benefited from the cooperation of both DS and single RRP strategies. The DS motion first tilts the rover's body toward the desired orientation, creating an angular momentum. Then, the single RRP induces more body turning by pedaling the right rear appendage while the remaining wheels keep doing DS motion. The continued DS motion helps the rover to agitate the material. reducing the material's resistance. Hence. the pedaling wheel can turn the rover more easily at each sweep-out, causing the body to pivot around the right rear appendage.

The Mini Rover experienced unsuccessful outcomes in 15 out of 30 iterations and 6 of these outcomes were right after the rover turned 90$^\circ$. This suggests that the center of mass (CoM) would be shifting outside the Rover's support polygon as its body orientation changed from one slope to another. Additionally, the Mini Rover's ability to generate reaction forces by pushing against the mound decreased a few seconds after the turning maneuver began. The sweeping leg cannot extend deep enough to reach the mound accumulated around the wheel. This is due to Mini Rover's inability to properly align its roll orientation and CoM position, to the changing slope conditions. Furthermore, in the single RRP motion, the wheel kept spinning during sweeping-out which limits the material resistance and creates low turning torque. 

\subsection{ML-inspired Gait Design}

We used these observations and gait cues obtained from BO-TRRP to update the gait scheme for fast turning. 
We first increased the DS speed of the wheels to further boost the initial downward tilting of the Mini Rover body. Then, we addressed the Mini Rover's roll angle adaptation issue by extending the wheels that are in the turning direction, facing the downward section of the slope, and retracting the wheels, facing the upward section of the slope. In doing so, the rover can adjust its body during and after the completion of its maneuver. This adjustment also enabled the Mini Rover to push against a larger granular mound and create high resistive force during sweep-out. 

To further enhance the turning behavior, we varied the spin and sweep speed parameters of the single RRP motion. We first, stopped the wheel spinning during the sweep-out motion and decreased the speed of the sweep, performing a solid manipulation on the substrate. In doing so, the wheel benefited more from the solid-like response of the mound surrounding the wheel by creating more sweeping torque in the direction of turning (Fig. \ref{Fig5_SoloWheel}A 1). Once the wheel started to sweep-in, we initiated the wheel spinning at a high rate and increased the sweep-in speed, generating a fluid manipulation (Fig. \ref{Fig5_SoloWheel}A 2)). This high-rate agitation created less torque during sweep-in, hence yielding less torque in the opposing direction to the Mini Rover turning. 

After these adjustments, we performed systematic maneuvering experiments with the new ML-inspired turning gait. The ML-inspired gait yielded a body turning of 90$^\circ$ in just above 4 seconds, outperforming previous gaits by a large margin. To compare the turning performance of the gaits presented in this study, we plotted the time evolution of body orientation trajectories of the Mini Rover in Fig. \ref{Fig4_PerfEval}. While TRRP achieved gradual and slow turning behavior, the optimized gaits initiated a fast turning, however, their performance degraded over time. The success of the ML-inspired gait lies in the combination of optimization and systematic human investigations. Understanding the locomotor-terrain interaction helps design more effective gaits, facilitating the Mini Rover to effectively manipulate its environment to maneuver its body.

\subsection{Locomotor-Terrain Interaction}
To understand and validate the existence of anisotropic torques created during single RRP, we performed single-wheel spin and sweep experiments to capture the resistive torques as illustrated with the images in Fig 6B. The blue curve in Fig. 6C illustrates the experiment where the wheel spins and sweeps at a higher rate, while the orange curve shows a non-spinning wheel and low-rate sweeping. The slow sweeping and non-spinning wheel induces a torque profile that is 2 times larger and 3 times longer than the fast sweeping and spinning wheel. Therefore, controlling the spin and sweep speed of the wheel enables creating different torque profiles. The findings in Fig. 6C support our experimental observations with Mini Rover that the wheel creates anisotropic torques by controlling the spin/sweep speeds. Hence, the directional torques facilitate rapid Mini Rover turning in the desired direction. 

\section{Conclusion}
In forthcoming extraterrestrial exploration missions, the ability to reach areas characterized by steep granular slopes and perform various maneuvers will be advantageous, serving both scientific data acquisition and mission navigation efficiency objectives. Nevertheless, current exploratory rover designs favor conventional wheels and passive suspension mechanisms that primarily allow navigation on flat granular surfaces with limited mobility over shallow granular slopes. To this end, by experimenting with a laboratory-scale rover robot capable of applying separate forces to each wheel system, we exploited the idea of manipulating granular terrain to achieve turning at steep granular slopes. 
This paper combines both machine learning and systematic investigation to obtain effective and robust turning gaits on granular materials. We first performed Bayesian Optimization to discover maneuvering gaits starting with a manually tuned turning gait. The optimization found two terrain manipulation strategies giving us deeper insight into how the Mini Rover achieved faster turning. Consequently, we performed systematic investigations and experiments using these strategies and further developed the optimized gaits. Hence, the combination of multiple techniques facilitated rover to turn 90$^\circ$ in about 4 seconds on a 25$^\circ$ granular slope. The Mini Rover achieved such fast turning by first differentially spinning its left and right-hand side wheels. Secondly, it creates isotropic torques with a single sweeping wheel that selectively spins and halts, creating a pivoting point for the rover. Finally, the rover adapts to the change of slope that occurs during turning by adjusting its body posture using extendible/retractable legs. 

The interaction between robots and terrain is crucial, particularly in deformable landscapes. Enhanced comprehension of this interaction through optimization and human intuition can contribute to the development of improved gaits and robots. The strategies shown in this study could prove valuable in designing maneuvering gaits for large-scale rovers and other exploration robots that are capable of terrain manipulation.

\bibliographystyle{IEEEtran}
\bibliography{root.bib}

\end{document}